\documentclass[runningheads]{llncs}

\usepackage{eccv}

\usepackage{eccvabbrv}

\usepackage{graphicx}
\usepackage{booktabs}
\usepackage{multirow}
\usepackage{makecell}
\usepackage{enumitem}
\usepackage{comment}
\usepackage{multirow}
\usepackage{makecell}
\usepackage{enumitem}

\usepackage{tabularx} 
\usepackage{float} 
\usepackage{colortbl}   
\definecolor{shadecolor}{rgb}{0.9, 0.9, 0.9} 
\usepackage[accsupp]{axessibility}  % Improves PDF readability for those with disabilities.

\usepackage{hyperref}

\usepackage{orcidlink}
\makeatletter
\newcommand{\printfnsymbol}[1]{%
  \textsuperscript{\@fnsymbol{#1}}%
}
\makeatother

\newcommand{\wu}[1]{\textcolor[rgb]{0,0,0} {#1}}

\newcommand{\mymodel}{WeCromCL}

\begin{document}

\title{WeCromCL: Weakly Supervised Cross-Modality Contrastive Learning for Transcription-only Supervised Text Spotting} 

\titlerunning{WeCromCL}

\author{Jingjing Wu\thanks{Authors contribute equally.}\inst{1}\orcidlink{0000-0002-1422-8984} \and
Zhengyao Fang\printfnsymbol{1}\inst{1}\orcidlink{0009-0000-0333-6210} \and
Pengyuan Lyu\printfnsymbol{1}\inst{2}\orcidlink{2222--3333-4444-5555} \and
Chengquan Zhang\inst{2} \and \orcidlink{0000-0001-8254-5773}
Fanglin Chen\inst{1} \and
Guangming Lu\inst{1} \orcidlink{0000-0003-1578-2634} \and
Wenjie Pei\textsuperscript{\dag}
\inst{1}\orcidlink{0000-0001-8117-2696}
}
\authorrunning{Jingjing Wu et al.}
\institute{Harbin Institute of Technology , Shenzhen, China 
\and
Department of Computer Vision Technology, Baidu Inc. \\
\email{\{jingjingwu\_hit, zhengyaonineve, wenjiecoder\}@outlook.com, lvpyuan@gmail.com, zhangchengquan@baidu.com, \{chenfanglin, luguangm\}@hit.edu.cn, }\\}


\maketitle
\renewcommand{\thefootnote}{\dag}
\footnotetext[2]{Corresponding author.}

\begin{abstract}
Transcription-only Supervised Text Spotting aims to learn text spotters relying only on transcriptions but no text boundaries for supervision, thus eliminating expensive boundary annotation. The crux of this task lies in locating each transcription in scene text images without location annotations. In this work, we formulate this challenging problem as a \textbf{We}akly Supervised \textbf{Cro}ss-\textbf{m}odality \textbf{C}ontrastive \textbf{L}earning problem, and design a simple yet effective model dubbed \textbf{\emph{\mymodel}} that is able to detect  each transcription in a scene image in a weakly supervised manner. % which associates each transcription with the correlated image region for effective text recognition. 
Unlike typical methods for cross-modality contrastive learning that focus on modeling the holistic semantic correlation between an entire image and a text description, our \emph{\mymodel} conducts atomistic contrastive learning to model the character-wise appearance consistency between a text transcription and its correlated region in a scene image to detect an anchor point for the transcription in a weakly supervised manner. The detected anchor points by \emph{\mymodel} are further used as pseudo location labels to guide the learning of text spotting. Extensive experiments on four challenging benchmarks demonstrate the superior performance of our model over other methods. Code is available at \url{https://github.com/ZhengyaoFang/WeCromCL}.%Moreover, we also evaluate our learned text spotter on the task of scene text retrieval without task-adaptation, which reveals that our method surprisingly compares favorably with other classical text spotters optimized by full supervision. 

\vspace{-5pt}

\keywords{Transcription-only supervised text spotting \and Weakly supervised cross-modality contrastive learning}

\end{abstract}  
%\vspace{-15pt}
\section{Introduction}
\label{sec:intro}
%\vspace{-4pt}
\begin{figure}[t]
%\vspace{-5pt}
  \centering
  \includegraphics[width=0.95\linewidth]{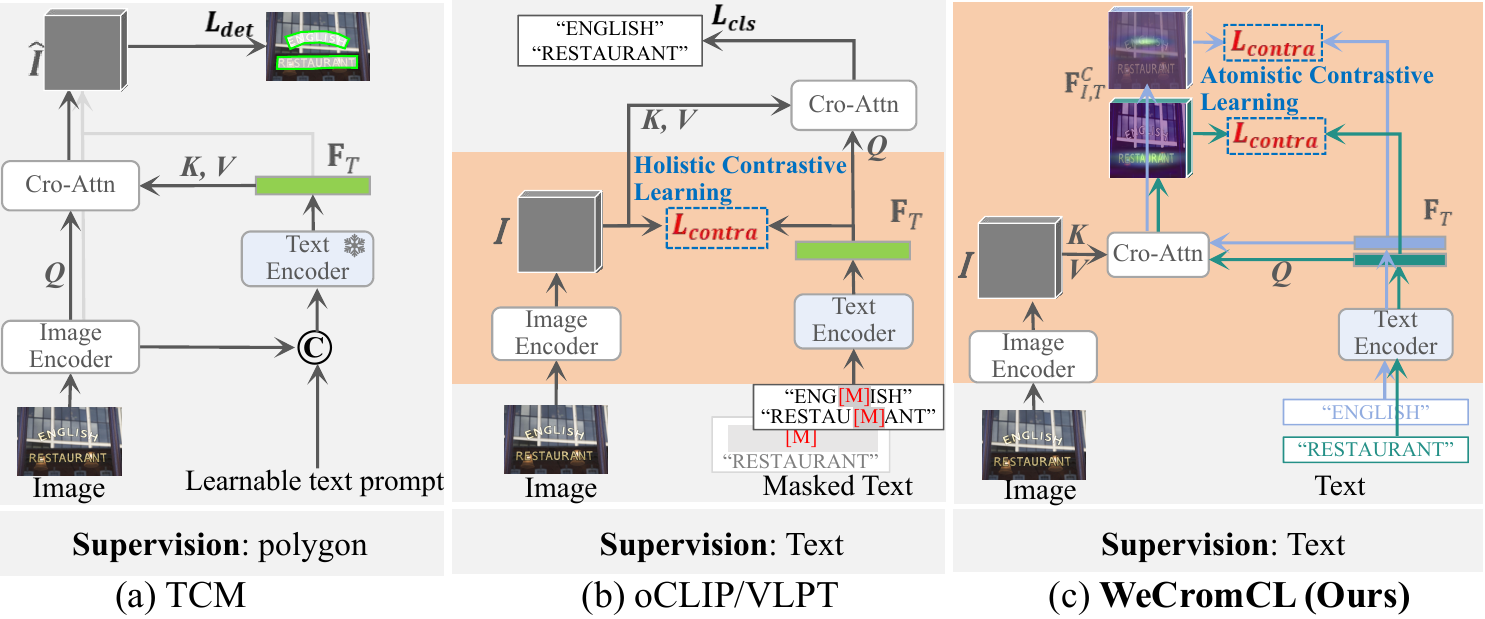}
  %\vspace{-10pt}
  \caption{
  Comparison between our \emph{WeCromCL} and TCM~\cite{Yu2023TurningAC}, oCLIP~\cite{Xue2022LanguageMA} as well as VLPT~\cite{song2022vision}. (a) TCM distinguishes text regions from non-text regions in a scene image in a fully supervised manner using text polygon annotations. (b) Both oCLIP and VLPT perform holistic contrastive learning between the entire scene image and the text in a full supervised way w.r.t. the contrastive pairs to learn effective image encoder for downstream OCR tasks, while relying on the auxiliary task for optimization, namely predicting masked characters (oCLIP) or masked words (VLPT). (c) Our \emph{\mymodel} conducts atomistic contrastive learning to model the appearance consistency between a text transcription and its correlated region in the scene image for transcription-wise detection in a weakly supervised manner without text location annotations.
  %The proposed \emph{\mymodel} performs weakly supervised cross-modality contrastive learning to learn the character-wise appearance consistency between a transcription and its correlated region in a scene image. The transcription acts as a cluster center that associates all positive images containing this transcription and the model is optimized to learn the similar appearance pattern for this transcription among all associated images, thereby leading to precise location of the transcription via learning the activation map without location annotations.
  }
  \label{fig:teaser}
  %\vspace{-25pt}
\end{figure}

%\vspace{-4pt}
Scene text spotting aims to detect and recognize text instances in scene text images. Existing methods~\cite{lyu2018mask,liao2019mask,liao2020mask,liu2020abcnet,liu2021abcnet,wang2020all,wang2021pan++,zhang2022text,wu2023single,wu2022decoupling,ye2023deepsolo,huang2023estextspotter} for text spotting have achieved remarkable progress relying on fully supervised learning, whereas these methods entail a large amount of annotations of text boundaries, which is extremely labor-consuming. In this work, we investigate transcription-only supervised text spotting, which only requires text transcriptions but no text boundaries for supervised learning, dramatically reducing the annotation overhead.

Transcription-only supervised text spotting is much more challenging than text spotting in full supervision, owing to the key difficulty of locating text transcriptions in scene text images without annotated text boundaries. A prominent method for transcription-only supervised text spotting is NPTS~\cite{peng2022spts}, which formulates the text spotting as a sequence prediction task. Specifically, it concatenates all text instances in a scene image into one sequence and seeks to predict all characters in an auto-regressive manner. While such modeling frees NPTS from text detection, a key limitation is that it suffers from arduous optimizing convergence. This is because there is no predefined order between different text instances when concatenating them together, the optimization of the model has to fit all potential permutations. Moreover, the model does not learn explicitly the mapping between text instances and correlated image regions without text detection, which further increases the difficulty of convergence. As pointed out as a primary limitation in the paper of NPTS, `the training procedure requires a large number of computing resources'. 
Another state-of-the-art method for transcription-only supervised text spotting is TOSS~\cite{tang2022you}, which draws inspiration from DETR~\cite{carion2020end} and locates text instances in scene images by pre-learning a set of text queries to probe transcriptions. However, the DETR-based methodology was initially designed for supervised object detection with location annotation. Although TOSS conducts modifications to adapt to weakly supervised text spotting, the absence of positional supervision still limits its effectiveness.%the effectiveness of spotting. 
%\vspace{-2pt}
%locate transcriptions accurately without annotated text boundaries for supervised learning.
% Another state-of-the-art method for Transcription-only supervised text spotting is TOSS~\cite{tang2022you}, which draws inspiration from DETR~\cite{carion2020end} and tries to locate text instances in scene images by pre-learning a set of text queries to probe transcriptions. However, such method depends heavily on the supervision of text boundaries, thus its effectiveness is limited without detection supervision. %locate transcriptions accurately without annotated text boundaries for supervised learning.

In this work we decompose the transcription-only supervised text spotting into two stages. % we propose to%resolve the task of transcription-only supervised text spotting by decomposing it into two stages. 
First, our method detects an anchor point for each transcription in a scene image to locate the correlated image region in a weakly supervised manner. Second, the obtained anchor points are used as pseudo location labels to learn a single-point supervised text spotter which relies on only one single point instead of text boundary as detection supervision.  %which associates each transcription with its correlated image region to ease subsequent text recognition. 
The first step, namely detecting the anchor points to locate transcriptions without the groundtruth, is particularly challenging, and meanwhile its detection accuracy is crucial to the performance of text spotting in the second stage. %Thus we focus on the first step. 
To address this problem, we formulate it as a weakly supervised cross-modality contrastive learning problem, and design a simple yet effective model dubbed \emph{\mymodel} for it. Unlike typical methods for cross-modality contrastive learning that focus on modeling the holistic semantic correlation between a text description and an entire image, as oCLIP~\cite{Xue2022LanguageMA} and VLPT~\cite{song2022vision} behave in Figure~\ref{fig:teaser}, our \emph{\mymodel} conduct atomistic contrastive learning to learn the character-wise appearance consistency between a text transcription and its correlated region in a scene image in a weakly supervised manner. In particular, we design a soft modeling mechanism to learn an activation map by measuring the appearance correlation between a transcription and each pixel of a scene image. The activated region in the scene image is identified as the anchor point for this transcription and is associated with the transcription for contrastive learning to optimize \emph{\mymodel}.

%Typical methods for weakly supervised detection include learning a spatial attention map based on solely visual features~\cite{xu2015show,pei2019attended,xu2016ask} or aggregating the feature maps contributed to one target category into an activation map~\cite{zhou2016learning}, both supervised indirectly by the final recognition loss. However, text spotting involves a huge number of categories when viewing each transcription as an individual category, which is extremely difficult for the above methods to perform effective learning guided by only recognition supervision. In contrast to these methods, we formulate the detection of anchor points for transcriptions as a Weakly Supervised Cross-Modality Contrastive Learning problem, and design a simple yet effective model dubbed \emph{\mymodel}. Unlike typical methods for cross-modality contrastive learning that focus on modeling the semantic correlation between an image and a text description, our \emph{\mymodel} aims to learn the character-wise appearance similarity between a text transcription and its correlated region around the anchor point in a scene image, as illustrated in Figure~\ref{fig:teaser}. In particular, we design a soft modeling mechanism to learn an activation map by measuring the appearance similarity between a transcription and each pixel of a scene image in a projected feature space. The activated region in the scene image is identified as the anchor point for this transcription and is associated with the transcription for contrastive learning to optimize \emph{\mymodel}.

Without the location annotations for text, our \emph{\mymodel} can still detect each transcription effectively. % an effective cross-modality similarity metric between a transcription and its correlated image regions by contrastive learning. 
The rationale is that a transcription acts as a cluster center that associates all matched images containing it and the model is optimized to learn the similar appearance pattern for this transcription among all associated images, leading to precise location of transcriptions via learning the activation map. %As shown in Figure~\ref{fig:teaser}, the detected anchor points by our \emph{\mymodel} are further used as pseudo location labels to guide the learning of a text spotter adapted from SRSTS~\cite{wu2022decoupling}, which performs text recognition relying on only one single anchor point rather than the precise text boundary for each transcription. 
To conclude, our contributions are summarized as follows:
\begin{itemize}[leftmargin =*, itemsep = 0pt, topsep = -2pt]
    \item We decompose the task of transcription-only supervised text spotting into two stages including weakly supervised text detection and single-point text spotting. Then we formulate the first and also challenging step as a cross-modality atomistic contrastive learning problem in an weakly supervised manner. 
    \item To the best of our knowledge, we are the first to define and investigate the weakly supervised atomistic contrastive learning problem between image and text modalities. We particularly propose a simple yet effective method for it, called \emph{\mymodel}, which can learn an effective cross-modality character-wise consistency metric between a transcription and its visual appearance in a scene image, thereby detecting the correlated image regions for the transcription without annotated text boundaries.
    \item Leveraging the predicted anchor points by our \emph{\mymodel} as pseudo location labels, we learn an effective single-point supervised text spotter adapted from SRSTS v2~\cite{wu2022decoupling, wu2023single}, a state-of-the-art text spotter. Integrating the proposed \emph{\mymodel} and the learned single-point text spotter, we construct a powerful system for transcription-only supervised text spotting, which compares favorably with existing methods on four challenging benchmarks. %In particular, we further evaluate our learned text spotter on the task of scene text retrieval without task-adaptation, which demonstrates that our text spotter even outperforms some classical text spotters by full supervision.
\end{itemize}

%\vspace{-3pt}
%\vspace{-5pt}
\section{Related Work}
%\subsection{Vision-Language Contrastive Learning}
%\vspace{-6pt}
\noindent\textbf{Vision-Language Contrastive Learning.} Vision-language contrastive learning has attracted increasing attention in recent years. A variety of vision-language contrastive learning methods~\cite{radford2021learning,jia2021scaling,li2021align,huo2021wenlan,he2020momentum,li2020unimo,yang2022vision,duan2022multi} are proposed for representation learning of both visual information and language prompt. These methods typically focus on learning the semantic correlations between text and image modalities, whereas our method aims to model the character-wise appearance similarity between a text transcription and its correlated region around the anchor point in a scene image. Recently, contrastive learning has also been introduced to OCR. A prominent example is oCLIP~\cite{xue2022language}, which conducts contrastive learning to optimize the image encoder for text spotting. It performs contrastive learning between an image and all the text instances appearing on the image in a holistic manner. Similarly, VLPT~\cite{song2022vision} also conducts holistic contrastive learning between an entire image and a transcription, and utilizes masked language modeling for auxiliary learning. Unlike oCLIP and VLPT, our \emph{\mymodel} seeks to learn the correlation between a transcription and the correlated image region for text location in a weakly supervised learning mode.

%\subsection{Scene Text Spotting}
% 1. seperate detector + recognizer;
% 2. end-to-end text spotting
% 3. arbitrary text spotting
\begin{comment}
    \noindent\textbf{Fully Supervised Text Spotting.}
The mainstream text spotters need precise boundaries for supervision. These methods can be generally divided into two categories: two-stage text spotters and one-stage text spotters. The typical two-stage methods~\cite{jaderberg2016reading,liao2017textboxes,liao2018textboxes++,lyu2018mask,liao2019mask,feng2019textdragon,liao2020mask,qiao2020text,wang2020all,liu2020abcnet,liu2021abcnet,wang2021pan++} regard text spotting as two sub-tasks: text detection and text recognition. They first conduct text detection under the supervision of  precise boundaries. To further perform recognition, various RoI pooling operations are introduced for local feature extraction.
Recently several one-stage methods are proposed to avoid RoI pooling. MANGO~\cite{qiao2020mango} regards text spotting as a pure text recognition task by a designed position-aware attention module. TESTR~\cite{zhang2022text} modifies Deformable DETR~\cite{zhudeformable} to deal with text spotting. SRSTS~\cite{wu2022decoupling} and SRSTS v2~\cite{wu2023single} decouple recognition from detection and propose a sampling-based text recognition mechanism. SPTS~\cite{peng2022spts} represents text instance as single point and tackles scene text spotting as a sequence prediction task. 
\end{comment}
\noindent\textbf{Fully Supervised Text Spotting.}
The mainstream text spotters need precise boundaries for supervision. 
The typical two-stage methods~\cite{jaderberg2016reading,liao2017textboxes,liao2018textboxes++,lyu2018mask,liao2019mask,feng2019textdragon,liao2020mask,qiao2020text,wang2020all,liu2020abcnet,liu2021abcnet,wang2021pan++} conduct detection and recognition serially and bridge them by RoI pooling operation. Recently several one-stage methods have been proposed. MANGO~\cite{qiao2020mango} regards text spotting as a pure text recognition task by a designed position-aware attention module. SRSTS~\cite{wu2022decoupling,wu2023single} decouples recognition from detection and proposes a sampling-based text recognition mechanism.  Several works ~\cite{zhang2022text, ye2023deepsolo, huang2023estextspotter} modify Deformable DETR~\cite{zhudeformable} to deal with text spotting. SPTS~\cite{peng2022spts} represents text instance as a single point and tackles scene text spotting as a sequence prediction task.

\noindent\textbf{Transcription-only Supervised Text Spotting.}
% 
% npts, toss, oCLIP, TSS ...
Currently, few works conduct text spotting under transcription-only supervision. Kittenplo \textit{et al.} ~\cite{kittenplon2022towards} refines Deformable DETR as an end-to-end text spotter named TTS. TTS is pre-trained on fully annotated synthetic data and fine-tuned on the transcription-only real-word data. It can be seen that TTS still uses a huge number of annotated synthetic data for training. Peng \textit{et al.}~\cite{peng2022spts} proposes no-point text spotting (NPTS) based on SPTS. NPTS takes transcription-only annotations as supervision and predicts randomly ordered transcriptions appearing in the scene text image. However, such design leads to arduous optimizing convergence and slow inference speed. TOSS~\cite{tang2022you} is transcription-only supervised and locates text instance by pre-learned queries, and its effectiveness is limited without detection supervision. Unlike the previous methods, we propose to conduct text spotting in two stages to ease transcription-only supervised text recognition problem: 1) detecting the anchor points for transcriptions; 2) conducting text spotting with the obtained anchor points as pseudo labels. As a result, our method circumvents the limitations suffered by previous transcription-only supervised methods.
%\vspace{-6pt}

%\vspace{-5pt}
\section{Method}
%\subsection{Overview}\vspace{-4pt}

\begin{figure*}[!t]
\centering
    \includegraphics[width=0.98\textwidth]{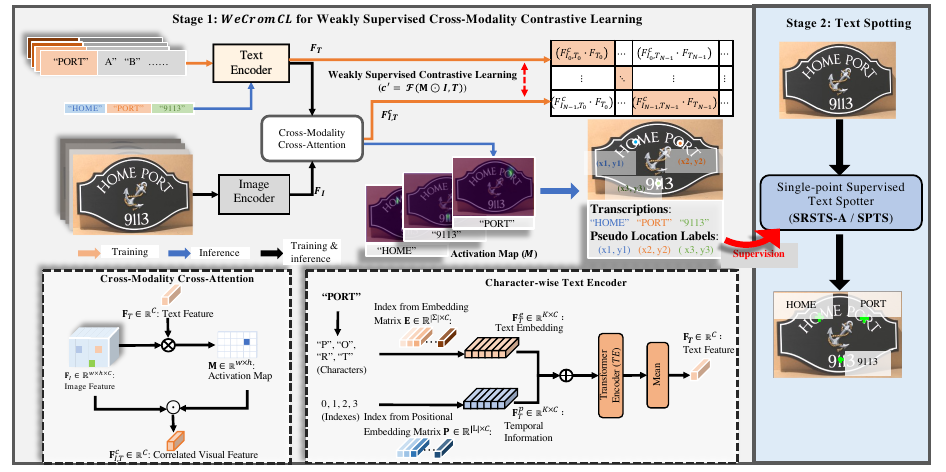}
\caption{Architecture of our proposed transcription-only supervised text spotter. Our method consists of two stages: 1) detecting the anchor point for each text instance as pseudo location label by \emph{\mymodel}; 2) conducting text spotting under the supervision of obtained pseudo location labels.
    }
\label{fig:Framework} 
\end{figure*}

Without annotations of text locations, it is difficult to apply the classical detect-and-recognize spotting paradigm~\cite{lyu2018mask,liao2019mask,liao2020mask,feng2019textdragon,liao2020mask,qiao2020text,wang2020all,liu2020abcnet,liu2021abcnet,wang2021pan++} to transcription-only supervised text spotting. %Thus, it is not feasible to apply the classical detect-and-recognize spotting paradigm~\cite{lyu2018mask,liao2019mask,liao2020mask,feng2019textdragon,liao2020mask,qiao2020text,wang2020all,liu2020abcnet,liu2021abcnet,wang2021pan++} that performs recognition relying on the precise detected boundaries. 
In light of this, we circumvent this difficulty by decomposing the task into two stages as shown in Figure~\ref{fig:Framework}: 1) detecting an anchor point in the scene image for each transcription to locate the correlated image region, and 2) leveraging the obtained anchor points as pseudo location labels to learn a single-point supervised text spotter which is learned based on only one single point as location annotation. The first step, namely detection of anchor points for transcriptions, is particularly challenging since the annotations of transcription locations are not available. Besides, The performance of text spotting in the second stage relies primarily on the predicting precision of the anchor points in the first stage. Thus, we focus on the first step and formulate it as a weakly supervised atomistic cross-modality contrastive learning problem, then we specifically design a simple yet effective framework dubbed \emph{\mymodel}.

\subsection{Weakly Supervised Atomistic Cross-Modality Contrastive Learning}
\label{sec:formulation}
%\smallskip\noindent\textbf{Problem Formulation.}
Typical cross-modality contrastive learning between text and image modalities, like CLIP~\cite{radford2021learning} or oCLIP~\cite{Xue2022LanguageMA}, %is defined to
\wu{aims to} learn the holistic semantic compatibility $c$ between an entire image $I$ and a text description $T$, which can be formulated as:
\begin{equation}
    c = \mathcal{F}(I, T),
    \label{eqn:typical-cl}
\end{equation}
where $\mathcal{F}$ denotes the transformation function of a contrastive learning model. In the task of transcription-only supervised text spotting, only the transcriptions contained in a scene image are provided whilst the annotation of text locations for each transcription is not available. Thus, we aim to estimate the location for each transcription to serve as the pseudo location labels for supervised learning of a text spotter. Formally, given an image $I$ containing a set of text transcriptions among which a text transcription $T$ is only associated with its corresponding region in $I$, the correlation $c'$ can be represented as:
\begin{equation}
    c' = \mathcal{F}(\mathbf{M} \odot I, T),
    \label{eqn:formulation}
\end{equation}
where $\mathbf{M}$ is an activation map whose size is equal to that of $I$ and $\odot$ denotes element-wise multiplication. All elements of $\mathbf{M}$ are  binary values indicating whether the corresponding pixel is associated with the transcription $T$. Since the groundtruth of $\mathbf{M}$ is not provided, we have to optimize the contrastive learning model $\mathcal{F}$ in a weakly supervised manner. Thus, we refer to such contrastive learning setting as weakly supervised atomistic contrastive learning between image and text modalities.

Formulating the detection of transcription in an image as the weakly supervised atomistic contrastive learning across modalities defined in Equation~\ref{eqn:formulation} involves two crucial challenges:
\begin{itemize}[leftmargin =*, itemsep = 0pt, topsep = -2pt]
    \item \emph{Challenge 1}: effective modeling of $\mathcal{F}$ entails precise estimation of the activation map $\mathbf{M}$ in weakly supervised learning without the groundtruth.
    \item \emph{Challenge 2}: unlike typical cross-modality contrastive learning such as CLIP that models the holistic semantic correlations between an entire image and a text, we aim to learn the atomistic correlation between a text transcription and its visual appearance in the correlated region in the scene image.
\end{itemize}  

To address these challenges, we design a simple yet effective model, namely \emph{\mymodel}, for weakly supervised atomistic contrastive learning. %explicated in Section~\ref{sec:mymodel}.

\begin{comment}
\smallskip\noindent\textbf{A Revisit to CLIP.}
CLIP~\cite{radford2021learning} is a prominent contrastive learning method for modeling the semantic correlation between an image and a text description, as shown in Equation~\ref{eqn:typical-cl}. 
CLIP cannot be directly applied to the detection of transcriptions in that it is unable to address the two challenges mentioned above. First, CLIP is designed in fully supervised mode rather than in weakly supervised mode, thus it cannot predict the activation map to detect the correlated region for a text transcription. Second, CLIP models the semantic correlations instead of character-wise appearance correlations between the text and image modalities.
\end{comment}

\subsection{\mymodel}

\label{sec:mymodel}
We propose \emph{\mymodel} to detect an anchor point for each text transcription to locate its correlated region in the scene image, which serves as the pseudo location label for optimizing the text spotter in the second stage. \emph{\mymodel} follows weakly supervised atomistic cross-modality contrastive learning framework. As formulated in Equation~\ref{eqn:formulation}, it takes a text transcription $T$ and an image $I$ as input, and predicts whether the image contains the transcription by measuring the correlation $c'$ between them. Meanwhile, \emph{\mymodel} predicts the activation map $\mathbf{M}$ in which the highly activated region corresponds to the associated image region for the text transcription and is identified as the anchor point.

As shown in Figure~\ref{fig:Framework}, similar to CLIP, \emph{\mymodel} employs an image encoder and a text encoder to learn latent embeddings for the input image and text transcription, respectively. In particular, we design a soft modeling mechanism to learn the activation map and thereby deal with \emph{Challenge 1}. Besides, we devise a character-wise text encoder for tackling \emph{Challenge 2}, which enables \emph{\mymodel} to learn the character-wise appearance similarity between the input transcription and its correlated region in the paired image. Finally, atomistic cross-modality contrastive learning is conducted to optimize the whole model of \emph{\mymodel}, using the constructed positive and negative training pairs based on the proposed negative-sampling mining scheme. %As a result, the proposed \emph{\mymodel} is able to address the two crucial challenges discussed in Section~\ref{sec:formulation}.

\noindent\textbf{Image Encoder.} %As shown in Figure~\ref{fig:Framework}, %we adopt the same model structure for the image encoder of \emph{\mymodel} in the first stage as that of the text spotter in the second stage to enhance the compatibility between two stages. Since our text spotter is adapted from SRSTS v2~\cite{wu2023single}, we directly inherit the structure of the image encoder from it. 
The image encoder of our \emph{\mymodel} first employs BiFPN~\cite{tan2020efficientdet} to extract multi-scale convolutional features, then enhances the feature learning by applying the deformable transformer encoder~\cite{zhudeformable}. The encoded image embeddings for the image $I$ are denoted as  $\mathbf{F}_I \in \mathbb{R}^{w\times h\times C}$.

\noindent\textbf{Character-Wise Text Encoder.} Typical cross-modality contrastive learning between image and text modalities, like CLIP, focuses on modeling the semantic correlation between two inputs. Thus, the text encoder of such models is designed to learn the holistic semantics of the input text. In contrast, our \emph{\mymodel} aims to learn the character-wise appearance consistency between the input text transcription and its visual appearance in the correlated image region. Thus, we devise the text encoder in the similar way as oCLIP~\cite{Xue2022LanguageMA} so that the encoded text embeddings 1) are distinguishable between different characters and 2) contain the temporal sequence information among characters in the text.

To learn text embeddings distinguishable between different characters, we learn individual vectorial embeddings with $C$ dimensions for each character in the alphabet $\Sigma$, which is equivalent to learning an embedding matrix $\mathbf{E} \in \mathbb{R}^{|\Sigma| \times C}$. Then we can encode the text transcription $T$ containing $K$ characters by indexing the corresponding embeddings from $\mathbf{E}$ for each character of $T$ sequentially and obtain the text embedding $\mathbf{F}_T^e \in \mathbb{R}^{K\times C}$.

To learn the temporal sequence information among characters in the text transcription, we learn extra positional embedding for each character position, resulting in an embedding matrix $\mathbf{P} \in \mathbb{R}^{L\times C}$ where $L$ indicates the maximum number of characters in a transcription. As a result, we can encode the temporal information $\mathbf{F}_T^p \in \mathbb{R}^{K\times C}$ for the transcription $T$ by indexing the positional embedding from $\mathbf{P}$ for all characters sequentially. We fuse the text embedding and the positional embedding by character-wise feature addition, and then adopt the Transformer encoder (TE) to perform feature propagation between characters in the transcription to model the correlation between them:

\begin{equation}
    \mathbf{F}_{T} = \text{Mean}(\text{TE}(\mathbf{F}_{T}^e + \mathbf{F}_{T}^p)),
\end{equation}
where $\mathbf{F}_{T} \in \mathbb{R}^{C}$ is the averaged text embedding over all characters by `$\text{Mean}$'.

%\vspace{-3pt}
\noindent\textbf{Soft Modeling of Activation Map by Cross-Modality Cross-Attention.} 
The key to estimating the activation map ($\mathbf{M}$ in Equation~\ref{eqn:formulation}) is how to measure the appearance correlation between the transcription and each pixel of the input image. To this end, we propose a soft modeling mechanism to learn such appearance correlation by measuring the cosine similarity between them in a projected feature space:

\begin{equation}
\begin{split}
    &\mathbf{M}_{(i,j)} = (\mathbf{W}_T^\top \mathbf{F}_T) \cdot  (\mathbf{W}_I^\top \mathbf{F}_{I,(i,j)}),\\
    &\mathbf{M} = \text{softmax}(\mathbf{M}).
    \label{eqn:soft}
\end{split}
%\vspace{-10pt}
\end{equation}
where $\mathbf{W}_T$ and $\mathbf{W}_I$ are learnable transformation matrices. $\mathbf{F}_{I,(i,j)}$ denotes the feature of pixel at $(i,j)$ in the image $I$.

The values of learned $\mathbf{M}$ are continuous values between $[0,1]$ instead of binary values while higher values indicate higher response to the transcription. Learning the activation map in such a soft modeling way eases the gradient propagation for optimization and can preserve richer similarity information than the hard representation by binary values. %Besides, the soft values of activation map reveal the consistency between the transcription and each pixel of the scene image while 
The most activated pixel with the peak value in the map can be identified as the anchor point for the transcription.

Following the formulation in Equation~\ref{eqn:formulation}, the learned activation map $\mathbf{M}$ is further used to aggregate the correlated features in the image to the text transcription for subsequent contrastive learning:

\begin{equation}
    \mathbf{F}_{I,T}^c = \sum_{i=0}^{w-1}\sum_{j=0}^{h-1} \mathbf{M}_{(i,j)} (\mathbf{W}_V^\top \mathbf{F}_{I,(i,j)}),
    \label{eqn:aggregation}
\end{equation}
where $\mathbf{F}_{I,T}^c \in \mathbb{R}^{C}$ is the aggregated correlated visual features in the image $I$ to the transcription $T$ and $\mathbf{W}_V$ is a learnable  matrix for feature transformation. Combining the soft modeling in Equation~\ref{eqn:soft} and the aggregation of correlated features in Equation~\ref{eqn:aggregation} essentially boils down to cross-modality cross-attention operation, where the encoded transcription feature $\mathbf{F}_T$ serves as the query while the all pixels of encoded image feature $\mathbf{F}_I$ serve as the keys and values.  

\smallskip\noindent\textbf{Cross-Modality Contrastive Learning by Negative-Sample Mining.} %or `negative-sample mining.'
We perform cross-modality contrastive learning between the learned correlated visual feature $\mathbf{F}_{I,T}^c$ and the encoded transcription feature $\mathbf{F}_T$ to optimize all modules of \emph{\mymodel} jointly. Similar to CLIP, for a positive training pair between an image $I$ and a transcription $T$, we construct negative pairs in two ways: either pair the image $I$ to multiple unpaired transcriptions (termed as image-to-text construction) or pair the transcription $T$ to multiple unpaired images (termed as text-to-image construction). 

%Following typical methods~\cite{radford2021learning,jia2021scaling,li2021align,huo2021wenlan} for contrastive learning, 
We maximize the Cosine similarity of positive pairs while minimizing the similarity of negative pairs. Formally, given a training batch of images $\{I_0, I_1,\ldots,I_{N-1}\}$ and their associated text transcriptions $\{T_0, T_1, \dots, T_{N-1}\}$, the loss function for the positive pair $(I_i, T_i)$ and negative pairs using the text-to-image construction is defined as:

\begin{equation}
    \mathcal{L}^{T2I}_{i} = - \text{log}\frac{\text{exp}(\text{Cosine}(\mathbf{F}_{I_i,T_i}^c, \mathbf{F}_{T_i})/\tau)}{\sum^{N-1}_{j=0} \text{exp}(\text{Cosine}(\mathbf{F}_{I_j,T_i}^c, \mathbf{F}_{T_i})/\tau)},
\end{equation}

Similarly, we can define the loss function for the positive pair $(I_i, T_i)$ and negative pairs using the image-to-text construction. In particular, we devise a negative-sample mining scheme to introduce more challenging negative pairs and thereby enhance the modeling robustness of \emph{\mymodel}. A straightforward way is to apply the hard-sample mining scheme that selects more similar but unpaired transcriptions with $I_i$ to construct more hard negative pairs. Surprisingly, we observe that randomly selecting unpaired transcriptions from the training set can also yield similar performance gain compared to hard-sample mining scheme as long as sufficient unpaired samples are provided. Thus, the loss based on such negative-sample mining scheme is defined as:
\begin{equation}
\mathcal{L}^{I2T}_{i} = - \text{log}\frac{\text{exp}(\text{Cosine}(\mathbf{F}_{I_i,T_i}^c, \mathbf{F}_{T_i})/\tau)}{\sum^{N+N_{\text{aug}}-1}_{j=0} \text{exp}(\text{Cosine}(\mathbf{F}_{I_i,T_j}^c, \mathbf{F}_{T_j})/\tau)},
\end{equation}
where $N_{\text{aug}}$ is the number of augmented negative pairs. Note that we only augment the unpaired transcriptions during image-to-text construction of negative pairs instead of augmenting the unpaired images during text-to-image construction, such negative-sample mining scheme can be performed quite efficiently with negligible overhead. Integrating the losses in two ways of negative pair construction, the loss of contrastive learning for a batch of N images is:
\begin{equation}
    \mathcal{L}_{\text{cm}} = \frac{1}{2N}\sum_{i=0}^{N-1}(\mathcal{L}^{T2I}_i + \mathcal{L}^{I2T}_i).
\label{eqn:cm}
\end{equation}

\begin{comment}
\begin{figure}[t]
    \centering
    \includegraphics[width=0.9\linewidth]{figures/cluster.pdf}
\caption{}
\vspace{-16pt} 
\label{fig:cluster} 
\end{figure}
\end{comment}

\noindent\textbf{Rationale behind \emph{\mymodel}.}
Our \emph{\mymodel} learns an effective cross-modality character-wise consistency metric between a transcription and the visual appearance in a scene image based on atomistic contrastive learning. It is able to detect the correlated region in the image to the transcription in a weakly supervised mode. The rationale behind this is that a transcription acts as a cluster center that associates all paired images with it, and the model is optimized to %encourage all associated images with this transcription to %have similar embeddings. Thus, the model is guided to 
learn the similar appearance pattern regarding this transcription among all the paired images to determine the activation map. Meanwhile, the optimization by minimizing the similarity between negative pairs can guide the model to learn discriminative appearance patterns for each transcription, thereby preventing the model from collapsing to a uniform pattern for different transcriptions. 
%\vspace{-6pt}

\subsection{Anchor-Guided Text Spotting}
%\pei{explain why \mymodel cannot perform text spotting directly.}
The most activated position in an activation map $\mathbf{M}$ learned from \emph{\mymodel} is identified as the anchor point for the corresponding transcription, which is further used as pseudo location label for learning the text spotter in the second stage. Theoretically, any existing single-point supervised text spotter can be readily applied to our framework. To validate the effectiveness of our \emph{\mymodel}, we conduct two instantiations of the text spotter in the second stage. We first instantiate it with SPTS~\cite{peng2022spts}, a prominent single-point supervised text spotter. Then we tailor a single-point text spotter specifically by adapting SRSTS v2~\cite{wu2022decoupling, wu2023single}, which is a state-of-the-art supervised text spotter, to construct an powerful transcription-only text spotting system. 

\noindent\textbf{Instantiation of Text Spotter with SPTS.} SPTS performs text spotting using only one single point for each transcription as location supervision. It formulates text spotting as a sequence prediction task. %Specifically, it concatenates all text instances in a scene image into one sequence and seeks to predict all characters in an auto-regressive manner. 
We use the predicted anchor points by our \emph{\mymodel} as pseudo location labels to train SPTS and its performance on text spotting can reflect the performance of \emph{\mymodel}.
%It first uses CNN followed by Transformer encoder to extract image features. The extracted features are further fed to Transformer decoder and predicted into token sequence that includes both detection and recognition results. SPTS only employs single point location and text annotation for training; thus, the pseudo-labels from \emph{\mymodel}  can naturally serve as supervision for SPTS.

\noindent\textbf{Instantiation by Adapting SRSTS.}
SRSTS v2 is initially designed using the text boundaries as location annotation for supervision, we adapt it to enable it to rely on only one single point during training and refer the adapted version as `SRSTS-A'. %and meanwhile further optimize several practical techniques involved in . 
We provide adaptation details including image encoding, model training and loss function design in the supplementary material. Integrating the proposed \emph{\mymodel} and SRSTS-A, we construct a powerful system for transcription-only supervised text spotting.

\vspace{-4pt}

%\vspace{-5pt}
%\vspace{-1pt} 
\section{Experiments}
%\vspace{-4pt} 
\subsection{Experimental Setup}
%\vspace{-5pt}
\noindent\textbf{Benchmarks.} 
1) ICDAR 2013~\cite{karatzas2013icdar} contains 229 training images and 223 testing images, in which most text instances are horizontal or slightly rotated. It provides ‘Strong’, ‘Weak’, and ‘Generic’ lexicons, which are represented as `S', `W' and `G' in Table~\ref{tab:spotting}. `S' denotes a lexicon containing 100 words, including the groundtruth transcription, which is provided for each test image. `W' means a lexicon that consists of all the words appearing in the test set. `G' is a generic lexicon provided by Liao \textit{et al.}\cite{liao2019mask}. 
2) ICDAR 2015~\cite{karatzas2015icdar} contains 1000 training images and 500 testing images. It involves oriented text instances annotated with quadrangles. 
3) Total-Text~\cite{ch2017total} comprises 1255 and 300 images for training and test, respectively. Most samples in this dataset are curved and are annotated with polygons and word-level transcriptions. `Full' lexicon is provided which includes all words in the testing set.
4) CTW1500~\cite{liu2019curved} consists of 1000 training images and 500 test images. The text instances are annotated at line-level and arbitrary-shaped. `Full' lexicon is provided for evaluation.

\noindent\textbf{Evaluation Protocol.}
Since our method only outputs transcriptions and corresponding anchor points, the evaluation protocol for fully supervised methods which relies on precise bounding box matching is not suitable for our method. We adopt the single-point and edit distance metrics, following SPTS~\cite{peng2022spts}.  For the single-point metric, we match each predicted anchor point with the nearest center point of groundtruth bounding boxes, and then check if their text content are consistent. As for the edit distance metric, matching is conducted solely based on the edit distance between predicted and groundtruth transcriptions.

\noindent\textbf{Implementation details.}
Following the previous methods~\cite{peng2022spts,liu2021abcnet,liu2020abcnet}, we train our method on a joint training set which consists of training images from %Synthtext~\cite{gupta2016synthetic},  
Curved Synthetic Dataset 150k ~\cite{liu2020abcnet}, ICDAR 2017 MLT~\cite{nayef2017icdar2017}, ICDAR 2013, ICDAR 2015 and Total-Text. In the first stage, we employ \emph{\mymodel} to generate the pseudo location labels for all training images. The obtained pseudo location labels are further used as supervision in the text spotting stage. Detailed settings are illustrated in the supplementary material.%~\ref{sec:appendix_sec1}. 
%\vspace{-8pt} 
\subsection{Ablation Studies of \mymodel}

\wu{In this section, we conduct ablation studies to investigate the effectiveness of proposed method. Note that more ablation studies and qualitative results are provided in the supplementary materials.}

\begin{table*}[!t]
   \centering
   %\vspace{-2pt}
  \caption{Ablation about different text encoders. `Token-wise' denotes the text encoder of CLIP focusing on learning the semantics of the entire text. `Character-wise' denotes the text encoder of \emph{\mymodel}. Single-point metric is used for evaluation. `P', `R' and `F' represent `Precision', `Recall' and `F-measure' respectively. } 
  %\vspace{-10pt}
  \label{tab:encoder}
  \scalebox{0.79}{\begin{tabular}{l|l|ccc|ccc|ccc|ccc}
  
   \toprule
    \multirow{2}{*}{Set}&\multirow{2}{*}{Text Encoder} & \multicolumn{3}{c|}{ICDAR 2013} &  \multicolumn{3}{c|}{ICDAR 2015}&  \multicolumn{3}{c|}{Total-Text} &  \multicolumn{3}{c}{CTW1500}\\
    \cmidrule(lr){3-14}
    & & P & R & F& P & R & F& P & R & F& P & R & F\\
  \hline
  \multirow{2}{*}{Training} &Token-wise &82.5 &82.5 &82.5 & 71.4 & 64.7 & 67.9 &68.1&69.3&	68.5 &53.1	&51.9&52.5\\
  & Character-wise &\textbf{93.2} &\textbf{93.2} &\textbf{93.2} & \textbf{91.4} & \textbf{86.0} & \textbf{88.6}& \textbf{83.5} & \textbf{85.1} & \textbf{84.3} & \textbf{67.0} & \textbf{65.7} & \textbf{66.3}\\
  \hline
  \multirow{2}{*}{Test}
  & Token-wise&79.7	&76.9&	78.6&65.7&63.3&64.4 &70.2&60.4&64.9&66.7&64.3&65.5\\
   &Character-wise& \textbf{90.4} &\textbf{90.5} &\textbf{90.5} & \textbf{86.9} & \textbf{80.1} & \textbf{83.4} & \textbf{85.8} & \textbf{75.4} & \textbf{80.3} & \textbf{78.9} & \textbf{76.5} & \textbf{77.7}\\
   %83.47	85.12	（76.5k）  84.29	85.66	75.41	80.21
  %& Character-wise&90.4&90.5&90.5&86.2&78.5&82.1&85.3&	74.5&66.7&64.3&65.5\\
  \bottomrule
 
\end{tabular}}
%%\vspace{-6pt}
\end{table*}

\begin{table*}[!t] 
  \caption{Ablation on the negative-sampling mining scheme for training \emph{\mymodel}.} %Negative-sampling mining scheme means more negative pairs are introduced to our cross-modality contrastive learning.} 
  %\vspace{-10pt} 
   \centering
  \label{tab:false}
  \scalebox{0.76}{\begin{tabular}{l|c|ccc|ccc|ccc|ccc}
   \toprule
    \multirow{2}{*}{Set}&\multirow{2}{*}{\makecell[l]{Negative-sampling \\mining scheme}} & \multicolumn{3}{c|}{ICDAR 2013} &  \multicolumn{3}{c|}{ICDAR 2015}&  \multicolumn{3}{c|}{Total-Text} &  \multicolumn{3}{c}{CTW1500}\\
    \cmidrule(lr){3-14}
    & & P & R & F& P & R & F& P & R & F& P & R & F\\
  \hline
   \multirow{2}{*}{Training} &$\times$ &91.5&91.5&	91.5&90.2&84.3&87.2&82.5&83.6&83.0&59.8&58.4&59.1\\
  & $\checkmark$ &\textbf{93.2} &\textbf{93.2} &\textbf{93.2} & \textbf{91.4} & \textbf{86.0} & \textbf{88.6}& \textbf{83.5} & \textbf{85.1} & \textbf{84.3} & \textbf{67.0} & \textbf{65.7} & \textbf{66.3}\\
  \hline
   \multirow{2}{*}{Test} &$\times$ &87.1&86.6	&86.8&86.2&78.5&82.1&85.3&74.5&79.5&68.4&65.9&67.1\\
  & $\checkmark$ &\textbf{90.4} &\textbf{90.5} &\textbf{90.5} & \textbf{86.9} & \textbf{80.1} & \textbf{83.4} & \textbf{85.8} & \textbf{75.4} & \textbf{80.3} & \textbf{78.9} & \textbf{76.5} & \textbf{77.7}\\
  
  \bottomrule
\end{tabular}}
%\vspace{-15pt}
\end{table*}

% \subsection{Transcription Detection by \mymodel}
%\emph{\mymodel} is designed for detecting the anchor point for each transcription. To quantify the performance of \emph{\mymodel}, we use the single-point metric for evaluation. In this section, we conduct ablation studies to demonstrate the effectiveness of the proposed components. In addition, quantitative and qualitative evaluation results are presented to validate the performance of \emph{\mymodel} for transcription detection. 
%\vspace{-5pt}
\noindent\textbf{Comparison between different text encoders.}
We further compare the character-wise text encoder of our \emph{\mymodel} with the token-wise text encoder of CLIP which focuses on learning semantics of the entire text. To be specific, we replace the text encoder of \emph{\mymodel} with the token-wise text encoder of CLIP and test its performance of transcription detection. We consistently use the prompt template ``There is a word `$transcription$'" for text encoding by CLIP, where `$transcription$' corresponds to the input text. %To bridge the distribution gap between the input of pre-trained CLIP and text transcription which only consists single or a few words in our task, we employ "There is a word \textbackslash "$label$\textbackslash"" as the prompt template, where $label$ will be replaced with corresponding text transcription. 

%The character-wise text encoder of our \emph{\mymodel} tokenizes the text transcription at the character level as input. In contrast, the widely used text encoder (token-wise) of CLIP employs a predefined language prompt and splits the input prompt as a sequence of tokens. To investigate the difference of the two types of text encoders, we conduct a comparative experiment by replacing our text encoder with the token-wise text encoder of CLIP. To bridge the distribution gap between the input of pre-trained CLIP and text transcription which only consists single or a few words in our task, we employ "There is a word \textbackslash "$label$\textbackslash"" as the prompt template, where $label$ will be replaced with corresponding text transcription. 

Table~\ref{tab:encoder} shows that %the performance of our \emph{\mymodel} equipped with two different text encoders respectively. %There is an obvious performance gap between the token-wise text encoder and our character-wise text encoder. 
%When equipped with the token-wise text encoder, our method performs poorly to query the position of text transcription. 
our model performs substantially better on all three metrics when equipped with the designed character-wise text encoder than using the token-wise encoder of CLIP. The results demonstrate that, compared to a token-wise text encoder that prioritizes semantic matching, encoding text at the character level can facilitate the learning of character-level correlations between a transcription and its visual representation in the image.

%Particularly, by employing the character-wise text encoder, the F-measure on the four benchmarks is improved by more than 10\%. 
% The results demonstrate that encoding transcriptions in a character-wise manner can facilitate the learning of character-wise correlation between a transcription and its visual appearance in the image.

\begin{table*}[!t] 
  \caption{Comparison between oCLIP using holistic contrastive learning and our proposed \emph{\mymodel} employing atomistic contrastive learning.}
  %\vspace{-8pt}
   \centering
  \label{tab:oclip}
  \scalebox{0.8}{\begin{tabular}{c|l|ccc|ccc|ccc|ccc}
   \toprule
    \multirow{2}{*}{Set}&\multirow{2}{*}{Methods} & \multicolumn{3}{c|}{ICDAR 2013} &  \multicolumn{3}{c|}{ICDAR 2015}&  \multicolumn{3}{c|}{Total-Text} &  \multicolumn{3}{c}{CTW1500}\\
    \cmidrule(lr){3-14}
    & & P & R & F& P & R & F& P & R & F& P & R & F\\
  \hline
   \multirow{2}{*}{Training} & oCLIP &87.0&79.9&83.3&51.1&35.4&41.9&70.4&43.5&53.8&45.2&37.4&40.9\\
  & \emph{\mymodel} &\textbf{93.2} &\textbf{93.2} &\textbf{93.2} & \textbf{91.4} & \textbf{86.0} & \textbf{88.6}& \textbf{83.5} & \textbf{85.1} & \textbf{84.3} & \textbf{67.0} & \textbf{65.7} & \textbf{66.3}\\
  \hline
   \multirow{2}{*}{Test} & oCLIP &76.6&68.9	&72.5&51.6&35.0&41.7&53.9&35.5&42.8&49.2&43.1&45.9\\
  & \emph{\mymodel} &\textbf{90.4} &\textbf{90.5} &\textbf{90.5} & \textbf{86.9} & \textbf{80.1} & \textbf{83.4} & \textbf{85.8} & \textbf{75.4} & \textbf{80.3} & \textbf{78.9} & \textbf{76.5} & \textbf{77.7}\\
  
  \bottomrule
\end{tabular}}
%\vspace{-8pt}
\end{table*}

% ic13 ic15 tt ctw
\noindent\textbf{Ablation on the negative-sample mining scheme.}
%We propose a negative-sample mining scheme by introducing more negative pairs in the training phrase of \emph{\mymodel}. 
To investigate the effectiveness of the proposed negative-sample mining scheme in \emph{\mymodel}, we compare the performance of two variants of \emph{\mymodel}: training with the negative-sample mining scheme and without the negative-sample mining scheme. The results in Table~\ref{tab:false} show that the negative-sample mining scheme yields large performance gain. Particularly, the F-measure on the training and test set of CTW1500 is improved by 7.2\% and 10.2\%, respectively. %indicatingmore negative pairs benefit the training of contrastive learning for the detection of transcriptions.%Since more challenge is introduced to the contrastive learning process, more precise localization performance is achieved. 

\begin{table}[!t]%\scalebox{0.7}
\centering
  \caption{Performance comparison between NPTS and \emph{\mymodel} + SPTS. The performance is evaluated by edit distance metric.}
  %\vspace{-10pt} 
   \label{tab:npts_vs_spts}
  \scalebox{0.8}{\begin{tabular}{l|ccc|cc}
   \toprule
   \multirow{2}{*}{Methods} &   \multicolumn{3}{c|}{ICDAR 2015 } & \multicolumn{2}{c}{Total-Text }\\
   \cmidrule(lr){2-6}
   & S & W & G & None & Full\\
   
  \hline
   NPTS & 70.3 & 62.7 & 57.0 & 61.6 & 70.6\\
   \emph{\mymodel} + SPTS &\textbf{71.8} & \textbf{64.7}& \textbf{59.7} & \textbf{63.2} & \textbf{70.7} \\
  \bottomrule
\end{tabular}}
%\vspace{-15pt}
\end{table}

\begin{figure}[t]
    \centering
    \includegraphics[width=0.55\linewidth]{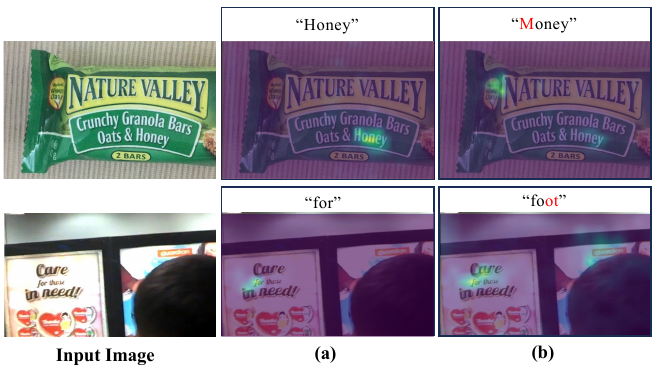}
    %\vspace{-10pt} 
\caption{Visual comparison of corresponding attention maps in the decoder of (a) \emph{\mymodel} + SPTS and (b) NPTS.}
%\vspace{-13pt} 
\label{fig:spts_vs_npts} 
\end{figure}

\noindent\textbf{Quantitative evaluation of transcription detection.} As shown in Table~\ref{tab:encoder} and Table~\ref{tab:false}, the obtained pseudo labels for training set are accurate and can serve as supervision for the following text spotting stage. Impressively, although contrastive learning is only conducted on the training set, \emph{\mymodel} still achieves generally satisfactory results on the test set. In particular, our \emph{\mymodel} achieves 90.5\%, 83.4\%, 80.3\% and 77.7\% on the test set of four benchmarks in terms of F-measure respectively. 

\noindent\textbf{Atomistic contrastive learning VS. holistic contrastive learning: comparison with oCLIP.} oCLIP performs holistic cross-modality contrastive learning between an entire scene image and all text appearing in the image to learn image encoder for OCR tasks. It employs an auxiliary task for optimization, which masks the characters for a transcription one by one and conducts prediction. We adapt it to weakly supervised text detection and compare it with our model. Specifically, we aggregate the predicted attention map for each masked character and the most activated pixel with the peak value in the aggregated map is identified as the location prediction for this transcription. Table ~\ref{tab:oclip} shows that our \emph{\mymodel}  outperforms oCLIP significantly on all benchmarks, which reveals the superiority of atomistic contrastive learning of \emph{\mymodel} over the holistic contrastive learning of oCLIP.     %is a weakly supervised pre-training method which designs a visual-textual decoder to model the interaction between each individual text instance and image. To quantify localization capability of oCLIP,  a systematic procedure is employed. Specifically, for a given input text, each character is successively masked, and the resulting attention map from the decoder is obtained. By aggregating peak points across all attention maps and calculating the mean, we obtain the pseudo location label for the input text. The performance comparison between \emph{\mymodel} and oCLIP is shown in Table ~\ref{tab:oclip}. As shown, our proposed \emph{\mymodel} impressively outperforms oCLIP on all benchmarks, which reveals the superiority of \emph{\mymodel} in modeling the relationship between individual text and the entire image.

\noindent\textbf{Comparison between \emph{\mymodel} + SPTS and NPTS.} As an indirect evaluation of our \emph{\mymodel}, we use the obtained pseudo location labels by \emph{\mymodel} to train SPTS and compare the performance with NPTS. \wu{SPTS is a prominent single-point text spotter while NPTS is its adapted transcription-only supervised variant. }%SPTS is a prominent  single-point  supervised text spotter while NPTS is the adapted transcription-only supervised version of SPTS.
\wu{For fairness, }%For a fair comparison, 
both NPTS and \emph{\mymodel} + SPTS are implemented based on their official code and neither utilize Random Cropping operation because it requires bounding box information. As shown in Table~\ref{tab:npts_vs_spts}, supervised by the pseudo labels from \emph{\mymodel}, \emph{\mymodel} + SPTS surpasses NPTS in all evaluation dimensions, particularly excelling in scenarios without the use of lexicons or use ‘Generic’ lexicon, where recognition accuracy is evident. Besides, the visualization results in Figure~\ref{fig:spts_vs_npts} also show the consistent results. These results reveal 1) the effectiveness of our \emph{\mymodel} for detecting transcriptions without annotations and 2) the superiority of the two-stage modeling strategy of our method over the single-stage method like NPTS.

\begin{figure}[t]
    %\vspace{-10pt} 
    \centering
    \includegraphics[width=0.5\linewidth]{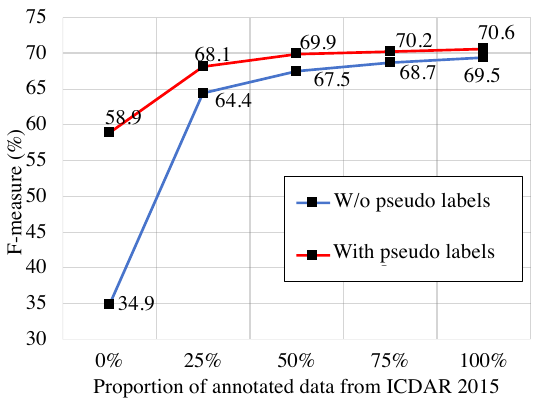}
    %\vspace{-10pt} 
\caption{Effectiveness of our proposed \emph{\mymodel} on full supervised spotting method. We pre-train SRSTS v2 on Curved Synthetic Dataset and fine-tune it on varying proportion of ICDAR 2015 gt and fixed amount of pseudo labeled data.}
%\vspace{-10pt} 
\label{fig:fully_supervised}
\end{figure}

\noindent\textbf{Enhancing Fully-Supervised Spotting.} Our \emph{\mymodel} can efficiently generate pseudo location labels from text-image pairs with no annotation cost. %Our \emph{\mymodel} can be used to generate pseudo location labels from transcription-image pairs efficiently with no annotation cost. 
Thus we can use it for pseudo data generation and investigate whether it can improve the performance of fully supervised single-point spotter. Specifically, we %utilize \emph{\mymodel} to 
generate pseudo labels for MLT, ICDAR 2013, Total-Text, and TextOCR~\cite{singh2021textocr}. Then we reproduce SRSTS v2 based on the Deepsolo framework~\cite{ye2023deepsolo} and pre-train it with Curved Synthetic Dataset. Note that SRSTS v2 can perform text recognition relying only on single point. During the fine-tuning stage, we train it with increasing annotated data from ICDAR 2015 and meanwhile evaluate the effect of adding fixed amount of (sufficient) pseudo-labeled data generated by \emph{\mymodel}. Figure~\ref{fig:fully_supervised} shows that \emph{\mymodel} can indeed improve the recognition performance of SRSTS v2, especially when the annotated data is not insufficient. Another interesting observation is that the performance of SRSTS v2 is quite limited when trained only on the synthetic data due to large data distribution gap between synthetic and real-world data. However, its performance is significantly improved when fine-tuned on the real-world data pseudo-labeled with our \emph{\mymodel}, while no human annotation cost is introduced.

\begin{comment}
    \begin{figure}[t]
\centering
    \includegraphics[width=0.6\linewidth]{figures/vis_attention_v3.pdf}
%\vspace{-10pt} 
\caption{Visualization of activation maps for predicting pseudo anchor points.
    }
    %\vspace{-18pt} 
\label{fig:pseudo_label}
\end{figure}
\end{comment}

\begin{figure}[t]
    \centering
    \includegraphics[width=1\linewidth]{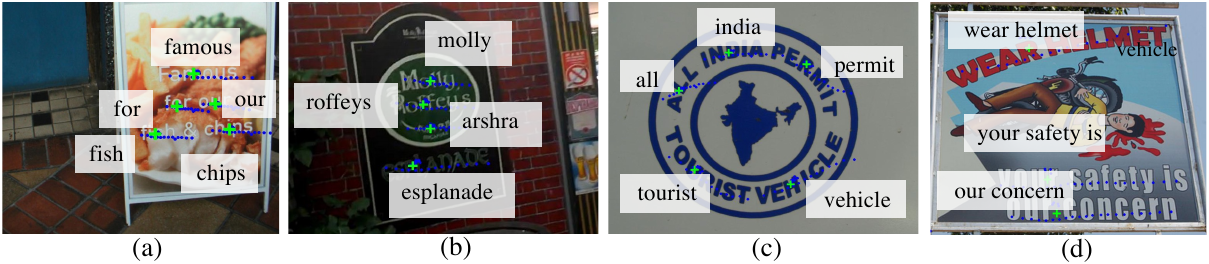}
    %\vspace{-17pt} 
\caption{Qualitative results of our transcription-only supervised text spotter on (a) ICDAR 2013, (b) ICDAR 2015, (c) Total-Text and (d) CTW1500. The \textcolor[RGB]{0,255,0}{green} `+' represents the estimated anchor point while \textcolor[RGB]{0,0,255}{blue} dots denote the sampled points.}
%\vspace{-20pt} 
\label{fig:spotter} 
\end{figure}

\begin{table*}[!t]
   \centering
   %\vspace{-5pt}
  \caption{Quantitative results on ICDAR 2013, ICDAR 2015, Total-Text and CTW1500. `*' denotes the performance evaluated by single-point metric. `$\dagger$' means the performance evaluated by edit-distance metric.} 
  %\vspace{-10pt}
  \label{tab:spotting}
  \scalebox{0.7}{
\begin{tabular}{l|ccc|ccc|cc|cc}
   \toprule
   \multirow{2}{*}{Methods} & \multicolumn{3}{c|}{ICDAR 2013} &  \multicolumn{3}{c|}{ICDAR 2015}&  \multicolumn{2}{c|}{Total-Text} &  \multicolumn{2}{c}{CTW1500}\\
   \cmidrule(lr){2-11}
   & S & W & G& S & W & G& None & Full & None & Full\\
  \midrule
  \multicolumn{11}{c}{Fully Supervised Methods}\\
  \midrule
  MTS v3~\cite{liao2020mask}&-&-&-& 83.3& 78.1& 74.2&71.2& 78.4&-&-\\
  MANGO~\cite{qiao2020mango} &93.4 &92.3 &88.7&85.4& 80.1& 73.9&72.9& 83.6&58.9 &78.7 \\

  ABCNet v2~\cite{liu2021abcnet}&-&-&-& 82.7& 78.5& 73.0&70.4& 78.1& 57.5& 77.2\\
  TESTR~\cite{zhang2022text}&-&-&-&85.2& 79.4& 73.6&73.3 & 83.9&56.0& 81.5\\
  %SRSTS v2~\cite{wu2023single}&-&-&-&{86.5}& {82.4}&{78.1}&{82.1}& {88.1}&61.2& 83.5\\
  TTS ~\cite{kittenplon2022towards} & - & - & - &85.2 &81.7& 77.4&78.2& 86.3& - & -\\
  ABINet++~\cite{fang2022abinet++} & - & - & -&  84.1 &80.4 &75.4 &  77.6 &84.5& 60.2& 80.3\\
  Deepsolo~\cite{ye2023deepsolo} & - & - & - &86.8 &81.9 &76.9 &  79.7 &87.0 &64.2 &81.4\\
  ESTextSpotter~\cite{huang2023estextspotter} &- & - & - &87.5 &83.0& 78.1& 80.8 &87.1& 64.9 &83.9\\
  SPTS$^{*}$~\cite{peng2022spts}&93.3 &91.7 &88.5 &77.5 & 70.2& 65.8&74.2 &82.4&{63.6}&{83.8}\\
  SPTS v2${*}$~\cite{liu2023spts} & 93.9& 91.8 &88.6& 82.3& 77.7& 72.6&75.5 &84.0& 63.6& 84.3 \\
  \midrule
  \multicolumn{11}{c}{Semi-supervised Methods}\\
  \hline
  TTS$_\text{weak}$ ~\cite{kittenplon2022towards}&- & - & - & 78.7 &75.2& 70.1 & 75.1 & 83.5 & - & -\\
  \hline
  \multicolumn{11}{c}{Transcription-only Supervised Methods}\\
  \midrule
  TOSS$^{*}$ ~\cite{tang2022you} & 86.4&85.1& 82.2 & 65.9&59.6&52.4&65.1&74.8&\textbf{54.2}&65.3\\
\textbf{\mymodel + SRSTS-A}$^{*}$ & \textbf{89.9}&\textbf{88.1}&\textbf{83.7}&\textbf{82.1}&\textbf{76.1}&\textbf{68.8}&\textbf{70.1}&\textbf{81.4}&51.2&\textbf{75.7}\\
 \midrule
  NPTS$\dagger$ ~\cite{peng2022spts} & 89.6 &86.4 & 83.2 & 70.3 &62.7 &57.0 & 61.6 & 70.6 &50.9 & 70.5
 \\
  \textbf{\mymodel + SRSTS-A} $\dagger$  &\textbf{91.2}& \textbf{89.8}& \textbf{84.6}& \textbf{79.5}& \textbf{72.8}& \textbf{66.2}& \textbf{68.1}& \textbf{79.1}& \textbf{52.7}& \textbf{79.9}\\

  \bottomrule
\end{tabular}
}
%\vspace{-15pt}
\end{table*}
\subsection{Transcription-only Supervised Text Spotting}
%\vspace{-6pt}
In this section, we evaluate our optimized system for transcription-only supervised text spotting, namely the integration of \emph{\mymodel} + SRSTS-A.   %performance of our two-stage transcription-only supervised text spotter with other spotting methods. To have a fair comparison, we report the performance evaluated by single-point metric and  edit distance metric respectively. Additionally, the qualitative results are shown in Figure~\ref{fig:spotter}.
\begin{comment}
    \wu{We noticed that NPTS employs the data augmentation operation of Random Cropping, which relies on bounding box information and  raises fairness concerns. Therefore, we removed this augmentation operation and reimplement NPTS based on their official code.}
\end{comment}

\noindent\textbf{Quantitative evaluation.} 
We evaluate our transcription-only spotting system on four benchmarks. For a fair comparison, we remove Random Cropping operation and re-train NPTS. As shown in Table~\ref{tab:spotting}, we achieve superior performance when compared with other transcription-only supervised methods. In particular, our method surpasses NPTS and TOSS by 16.4\% and 9.2\% on ICDAR 2015 when evaluated with generic lexicon. Our method also performs well on the challenging Total-Text with curved text. Supervised by \emph{\mymodel}, SRSTS-A impressively outperforms TOSS by 5\% and NPTS by 6.5\% in the metric of `None'. 

TTS$_\text{weak}$~\cite{kittenplon2022towards} is a semi-supervised text spotter which follows DETR-based spotting framework. It is trained on fully-annotated synthetic data (including annotations of text boundaries) and transcription-annotated real-world data. As shown in Table~\ref{tab:spotting}, our method still achieves comparable performance compared to TTS$_\text{weak}$ although our spotting system only uses pseudo location labels generated by our \emph{\mymodel} for all data.%The results demonstrate the effectiveness of our optimized supervised text spotting system and the robustness of proposed \emph{\mymodel}.

\noindent\textbf{Qualitative evaluation.}
We visualize the spotting results in Figure~\ref{fig:spotter}. %More visualizations are provided in the supplementary material. 
As shown, our optimized system can handle various challenging scenarios, like tiny, fuzzy, curved  and long text. The visualization results indirectly indicate the effectiveness and robustness of  \emph{\mymodel}.
\section{Conclusion}
%\vspace{-2pt}
%\vspace{-2pt}
In this work, we decompose the transcription-only supervised text spotting into two stages: 1) detecting the anchor point for each transcription and 2) conducting text spotting guided by the obtained anchor points, among which the first stage is quite challenging and crucial. We formulate the detection of anchor points for text transcriptions as a weakly supervised atomistic contrastive learning problem across modalities, and devise a simple yet effective method dubbed \emph{\mymodel} for it. %Our \emph{\mymodel} learns the character-wise appearance similarity between a text transcription and its correlated image region in a weakly supervised manner. 
The detected anchor points are further used to guide the learning of text spotting. Extensive experiments on challenging benchmarks demonstrate the effectiveness and advantages of our proposed method. %In addition, our method also achieves impressive results on the scene text retrieval task, which further reveals the superior performance of our model .

\noindent\textbf{Limitations.} %Due to the fact that small texts are only discernible on high-resolution input images, \emph{\mymodel} requires a larger input size to ensure proper recall of small texts, which limits the speed of inference in obtaining pseudo labels.
To measure the character-wise appearance consistency accurately between a transcription and the correlated region in the scene image, our \emph{\mymodel} %typically
requires high resolution of input images for detecting small texts. 

\section*{Acknowledgements}
This work was supported in part by the National Natural Science Foundation of China (U2013210, 62372133),  in part by Shenzhen Fundamental Research Program (Grant NO. JCYJ20220818102415032), in part by Guangdong Basic and Applied Basic Research Foundation (2024A1515011706), in part by the Shenzhen Key Technical Project (NO. JSGG20220831092805009, JSGG20201103153802006, KJZD20230923115117033).

\clearpage  % TODO FINAL: This \clearpage needs to be removed from both review and camera-ready versions.


\bibliographystyle{splncs04}
\bibliography{main}
\title{Supplementary of \mymodel}

\maketitle

\section{Instantiation by Adapting SRSTS}
SRSTS-A consists of Image Encoder, Anchor Estimator, Sampling Module and Recognition Module, which is consistent with SRSTS. Taking extracted features from Image Encoder, Anchor Estimator predicts the anchor point for each text transcription. Meanwhile, Sampling Module performs sampling around each anchor and provides Recognition Module with the sampled features for text decoding. While the modeling details of SRSTS can be found in the corresponding paper, we elaborate on the differences between it and the adapted SRSTS-A. 

We adapt SRSTS to SRSTS-A with three major modifications. First, the detection branch of SRSTS using text boundaries is dropped in our text spotter. Second, we incorporate several practical techniques of DeepSolo (ResNet-50) ~\cite{ye2023deepsolo} into SRSTS-A for system enhancement, including the image encoder, data augmentation and optimizating strategies. Third, we modify the loss function for Anchor Estimator $\mathcal{L}_c$ of SRSTS from Dice loss~\cite{milletari2016v} to Focal loss~\cite{lin2017focal} to adapt to the supervision variation from text boundaries to anchor points for better convergence. To be specific, given a feature map $\mathbf{P}$, Anchor Estimator of SRSTS-A learns a confidence map to indicate the probability of each pixel to be an anchor. It employs a 1$\times$1 convolutional layer followed by Sigmoid function to generate the confidence map $\mathbf{C}$. We use Focal loss to optimize the parameters of Anchor Estimator:
\vspace{-4pt}
\begin{equation}
\vspace{-4pt}
\begin{split}
    &\mathcal{L}_c = \sum_{i=0}^{w-1}\sum_{j=0}^{h-1} -\alpha\mathbf{C}_{\text{gt}}(i,j)  (1-\mathbf{C}(i,j))^\gamma \text{log} (\mathbf{C}(i,j)) \\
    &-(1-\alpha)(1-\mathbf{C}_{\text{gt}}(i,j)) \mathbf{C}(i,j)^{\gamma} \text{log} (1-\mathbf{C}(i,j)),\\
\end{split}
\end{equation}
where $\alpha$ and $\gamma$ are weighting factors for focal loss. $\mathbf{C}_{\text{gt}}$ is pseudo groundtruth for the confidence map constructed from the obtained anchor point by \emph{\mymodel}: the anchor point is assigned 1 and other pixels are assigned 0. 

Integrating \emph{\mymodel} and SRSTS-A, we obtain the optimized system for transcription-only supervised text spotting.

\section{Implementation Details}
\noindent\textbf{Implementation details of \emph{\mymodel}.}
We firstly pre-train  \emph{\mymodel} on synthetic datasets (Synthtext~\cite{gupta2016synthetic} and Curved Synthetic Dataset) for 200,000 steps with batch size set to be 16. The input size is set to be (640, 640) for fast convergence. Then it will be fine-tuned on the training set of each dataset for 80,000 steps respectively with batch size set to be 4. The following data augmentation strategies are conducted during training: 1) randomly resize the short side of the input image to a range from 640 to 896 while keeping the longer side shorter than 1,280; 2) randomly rotate the input image; 3) randomly apply blur and color jitter. 
Our method is optimized by SGD with initial learning rate 1e-3 on synthetic datasets and 1e-4 on specific real word dataset. The weight decay is set to be 0.0001 and momentum is set to be 0.9. The learning rate is delayed with a ‘poly’ strategy. When inferring images to obtain the pseudo location labels, we resize the longer side of input image for ICDAR 2013, ICDAR 2015 to 1152 and 1696, and the shorter side of input image for Total-Text and CTW1500 to 896 and 992. 

\noindent\textbf{Implementation details of spotting.}
In the text spotting stage, our text spotter is supervised by the obtained pseudo location labels. The text spotter is pre-trained on the joint training dataset that contains Curved Synthetic Dataset, ICDAR 2017 MLT, ICDAR 2013, ICDAR 2015, and Total-Text with pseudo location labels for 425,000 steps at first. For word-level benchmarks, our text spotter is fine-tuned on the training set of specific benchmark for 3,000 steps.  For CTW 1500, we use line-level text transcriptions of SynthText~\cite{gupta2016synthetic} to generate line-level pseudo location labels and further train the text spotter for 100,000 steps based on the obtained pseudo line-level location labels. Finally, the pre-trained model is further fine-tuned on CTW 1500 training set for 20,000 steps. Adam is used as optimizer. The learning rate is set the same as Deepsolo, and the same data augmentation is used except for Random Cropping operation being removed.
%We employ SGD as optimizer and set the initial learning rate to be 1e-4. The learning rate is delayed with `poly' strategy. The weight decay and momentum are set to be 0.0001 and 0.9 respectively. 
In the testing phrase, we resize the shorter side of input image to 864, 864, 1440 and 576 for ICDAR 2013, ICDAR 2015, Total-Text and CTW1500 respectively.

\section{Ablation Studies of \mymodel}

\begin{table}[!ht]
    \centering
    \caption{Comparison of Pseudo Label Quality on ICDAR 2015.}
    \label{tab:pseudo}
    \small
    \scalebox{1}{
\begin{tabular}{c|l|c|c|c}
        \toprule
        Set & Method &P & R & F \\
        \hline
        \multirow{3}{*}{Training} &oCLIP & 51.1 & 35.4 & 41.9 \\
        &NPTS & 47.6 & 48.2 & 47.9\\
        & \emph{\mymodel} & \textbf{91.4}& \textbf{86.0} &\textbf{88.6}  \\
        % \hline
        % \multirow{3}{*}{Test} &NPTS& 39.0 & 33.6 & 35.7 \\
        % &oCLIP & 51.6 & 35.0 & 41.7 \\
        % & \emph{\mymodel} & \textbf{86.9} &\textbf{80.1}& \textbf{83.4}\\
        \bottomrule
    \end{tabular}}
\end{table}
\begin{table}[t]
    \centering
    \captionof{table}{Spotting results under different pseudo-label generation methods and ablation conditions on ICDAR 2015. `gt` and `pse` indicates actual ground truth and pseudo labels, respectively. Subscripts `trans`, `point`, and `box` specify text-only, single-point, and bounding box annotations, respectively. Notably, `pse$_\text{point}$` utilizes the same text-only annotation information as `gt$_\text{trans}$`. }
    \label{tab:spotter}
    \scalebox{1}{\begin{tabular}{c|l|c|c|c|c|c|c}
        \toprule
        \multirow{2}{*}{Row}&\multirow{2}{*}{Method}&\multicolumn{2}{c|}{Data}&\multirow{2}{*}{$\alpha$}&\multirow{2}{*}{S}&\multirow{2}{*}{W} & \multirow{2}{*}{G} \\
        \cmidrule(lr){3-4}
        & & Synth & Real & & & &\\
        \hline
        1&oCLIP& pse$_\text{point}$-150k & pse$_\text{point}$-11k & $-$ & 60.9 & 57.7 & 51.7\\ 
        2&NPTS& pse$_\text{point}$-150k  &  pse$_\text{point}$-11k & $-$ & 72.7 & 68.4 & 61.7\\
        %\hline
        %3&TTS$_\text{weak}^*$& gt$_\text{box}$-800k & gt$_\text{trans}$-43k & $-$ & 78.7 & 75.2 & 70.1\\
        %4&Ours& pse$_\text{point}$-800k & pse$_\text{point}$-43k & $-$ & 82.4 & 77.5 & 70.2\\
        \hline
        \rowcolor{shadecolor}
        3& Ours & pse$_\text{point}$-150k & pse$_\text{point}$-11k & $-$ &82.1 &76.1 &68.8\\
        4& Ours & gt$_\text{point}$-150k & pse$_\text{point}$-11k & $-$ &82.8 & 76.3 & 69.4\\
        5& Ours & gt$_\text{point}$-150k & gt$_\text{point}$-11k & $-$ &84.8 & 78.4 & 71.4\\
        6& Ours & gt$_\text{box}$-150k & gt$_\text{box}$-11k & $-$ & \textbf{86.8} & \textbf{82.4} & \textbf{77.8} \\
        \hline
        7& Ours & pse$_\text{point}$-150k & gt$_\text{point}$-11k & 0 & 82.9 & 77.0 & 69.8\\
        8& Ours & pse$_\text{point}$-150k & gt$_\text{point}$-11k & 0.3 & 79.1 & 74.6 & 68.0\\
        9& Ours & pse$_\text{point}$-150k & gt$_\text{point}$-11k & 1 & 42.0  & 39.9 & 36.2\\
        \bottomrule
    \end{tabular}}
\end{table}
\noindent\textbf{Comparison with Other Keypoint Localization / Pseudo-labeling Methods.} To the best of our knowledge, our approach is the first to generate localization pseudo labels using text-only supervision. Our approach’s efficacy is evidenced by comparing our pseudo label generation method with those derived from the attention map of another weak supervision method, oCLIP, as presented in Table 3 of our submission. Here, we further include another weak supervision method, NPTS, and evaluate the impact of pseudo labels generated by different methods on the final spotting performance. %
\wu{The experimental results are presented in Table~\ref{tab:pseudo} and rows 1, 2, and 3 of Table~\ref{tab:spotter}, confirming the effectiveness of our method.}

\noindent\textbf{Impact of Anchor Quality on Spotting Results.} In Table~\ref{tab:spotter}, we compare our method's accuracy (row 3) to ground truth anchor points (row 7) and perturbed anchor points (rows 8, 9) in the spotting task. With $\alpha$ representing the perturbation degree, where $\alpha$ is set to 0.3, the offset from ground truth follows a Gaussian distribution within 0.3 times the text box size. Results show our method's accuracy is close to using ground truth (82.1/76.1/68.8 vs 82.9/77.0/69.8), confirming the accuracy of our pseudo labels. \wu{In addition, we trained with ground truth from synthetic data and pseudo label from real-word data (row 6 in Table~\ref{tab:spotter}, which does result in a slight improvement. However, this slight difference underscores: 1) the high quality of our method's pseudo labels; 2) our method's effectiveness without relying on explicit position labels. } 

\noindent\textbf{Overall Gap between Fully Supervised Methods.} In Table~\ref{tab:spotter}, we observe that the performance gap primarily stems from two factors: pseudo label accuracy (rows 3, 4, 5) and the positional supervision method (rows 5, 6). Comparing ground truth point supervision to pseudo labels supervision, the performance gaps are 2.7, 2.3, and 2.6, respectively. Similarly, the gaps between box supervision and point supervision are 2, 4, and 6.4, highlighting the importance of detailed positional information. Moving forward, we aim to explore methods for obtaining high-quality pseudo labels at the box level.

\section{Visualization Results.}

\begin{figure}[]
\centering   
\includegraphics[width=0.98\linewidth]{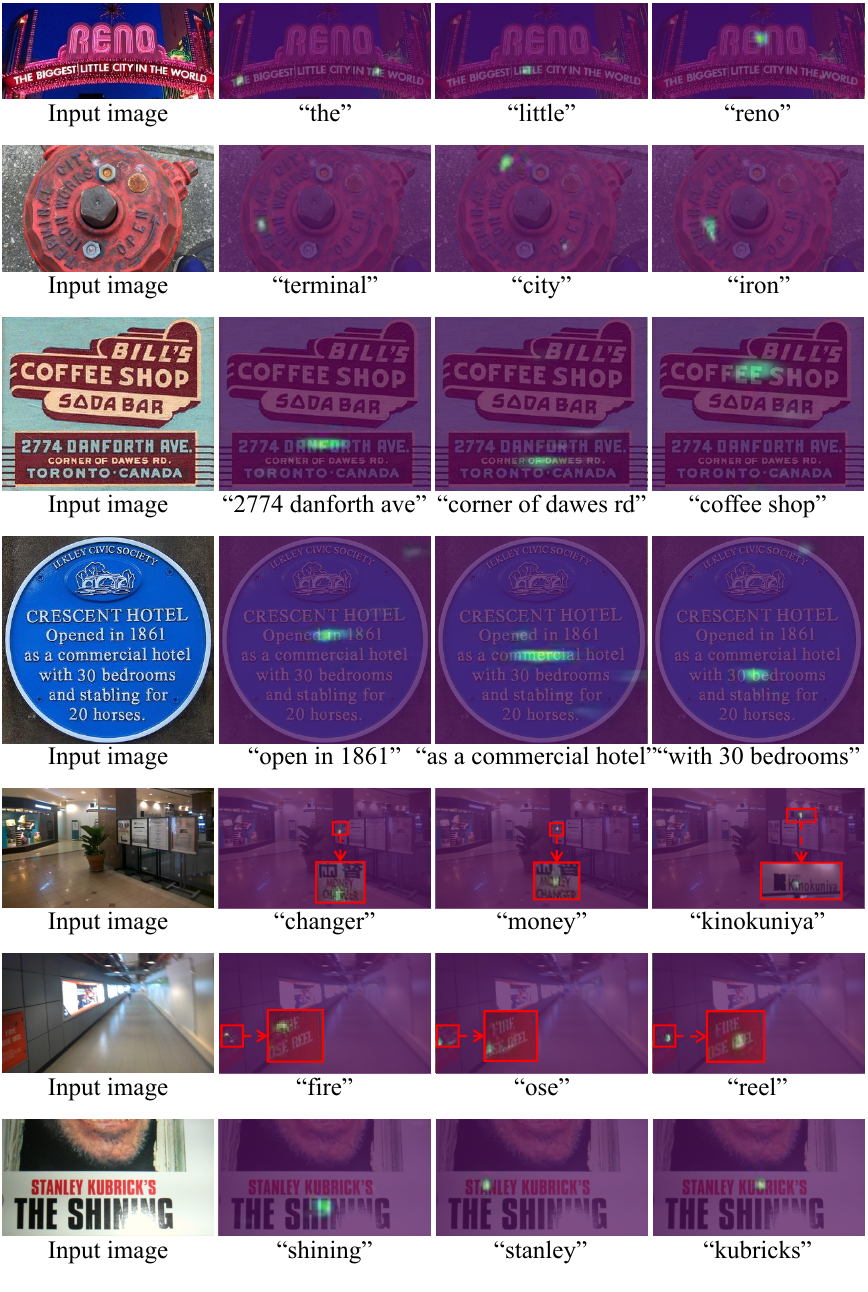}
\vspace{-20pt}
\caption{Visualization of activation maps learned by \emph{\mymodel}.  Our \emph{\mymodel} can handle various complex cases, such as text with artistic fonts, curved text, long text, and small text. Given a text transcription,  \emph{\mymodel} can generate corresponding activation map in which the highly activated region is identified as the anchor point for this transcription.}
\vspace{-15pt}
\label{fig:supp_1} 
\end{figure} 

\begin{figure*}[t]
\centering   
\includegraphics[width=1\linewidth]{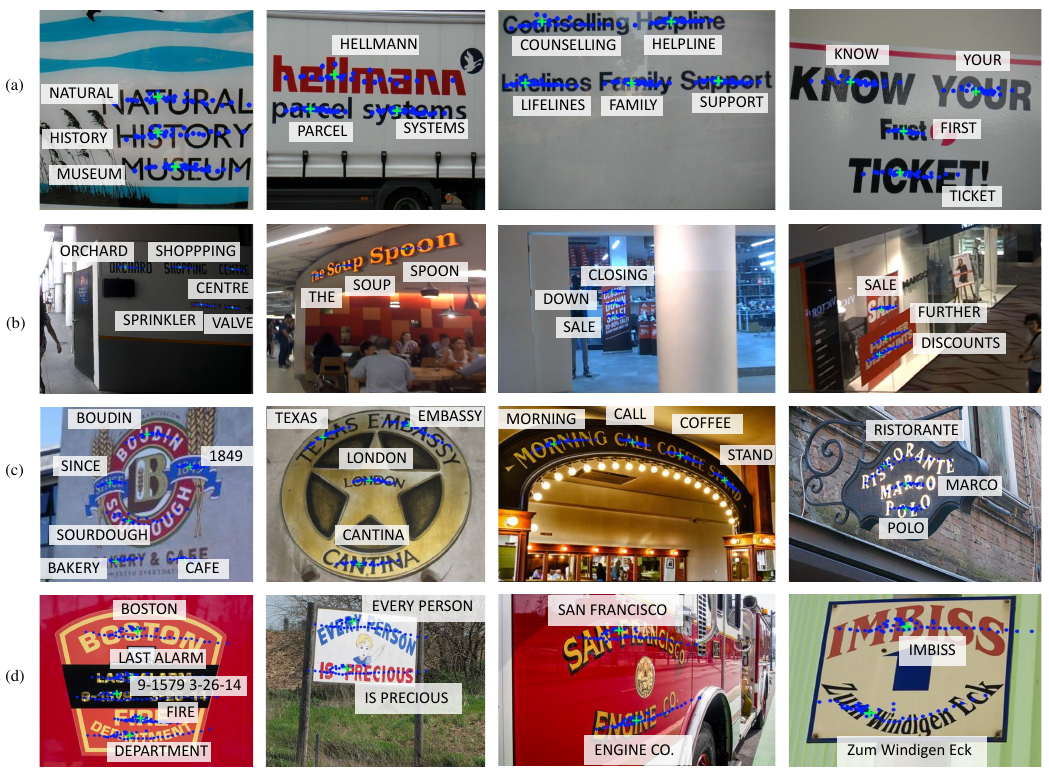}
\vspace{-16pt}
\caption{Visualization of text spotting results on four benchmarks: (a) ICDAR 2013, (b) ICDAR 2015, (c) Total-Text and (d) CTW1500. The \textcolor[RGB]{0,255,0}{green} `+' represents the estimated anchor point for each text instance. The \textcolor[RGB]{0,0,255}{blue} dots denote the sampled points.
    }
\vspace{-15pt}
\label{fig:supp_2} 
\end{figure*} 

\subsection{Visualization of Activation Maps}
\vspace{-4pt}
To better illustrate the localization performance of \emph{\mymodel}, we show sufficient activation maps generated by \emph{\mymodel} in Figure~\ref{fig:supp_1}. We can observe that our \emph{\mymodel} successfully locates the text region when given a text query. Even when the queried text is small and fuzzy within the image, by enlarging the input image, \emph{\mymodel} is still capable of successfully locating the most relevant position associated with the queried text. The most activated pixel with the peak value in each activation map is identified as the anchor point for the corresponding transcription. The obtained anchor points are further used as pseudo location labels to supervise the learning of text spotter in the text spotting stage.

\subsection{Visualization of Text Spotting Results}
\vspace{-4pt}
Some text spotting results are shown in Figure~\ref{fig:supp_2}. Our text spotter is learned under the supervision of pseudo location labels obtained by \emph{\mymodel}. As can be easily seen, the proposed transcription-only supervised text spotter can achieve satisfactory performance even when facing challenging cases such as tiny text, fuzzy text, curved text and long text. With the provided precise pseudo location labels as supervision, our text spotter learns to locate the text instance precisely and successfully performs sampling for text recognition. The visualization of text spotting results intuitively demonstrate the effectiveness and robustness of proposed transcription-only supervised text spotter.

\end{document}